\documentclass[10pt,twocolumn,letterpaper]{article}

\usepackage{iccv}
\usepackage{amsmath,graphicx}
\usepackage{times}
\usepackage{epsfig}
\usepackage{amssymb}
\usepackage{nicefrac}
\usepackage{epstopdf}
\usepackage{booktabs}
\usepackage{multirow}
\usepackage{subcaption}
\usepackage{dblfloatfix}
\usepackage[linesnumbered,ruled]{algorithm2e}
\usepackage{algpseudocode}
\usepackage{bm}
\usepackage{dsfont}
\usepackage[pagebackref=true,breaklinks=true,letterpaper=true,colorlinks,bookmarks=false]{hyperref}
\usepackage{array}
\newcolumntype{P}[1]{>{\centering\arraybackslash}p{#1}}

\iccvfinalcopy 

\def\iccvPaperID{1} 

\begin{document}

\title{Curriculum Learning for Multi-Task Classification of Visual Attributes}

	\author{Nikolaos Sarafianos$^{1}$ \\
		Christophoros Nikou$^{3}$
		\and
		Theodore Giannakopoulos$^{2}$\\
		Ioannis A. Kakadiaris$^{1}$
		\and		
	$^{1}$University of Houston \quad $^{2}$NCSR Demokritos \quad $^{3}$University of Ioannina
	}
	
	\maketitle   
\thispagestyle{empty}

\begin{abstract}
   	Visual attributes, from simple objects (\eg, backpacks, hats) to soft-biometrics (\eg, gender, height, clothing) have proven to be a powerful representational approach for many applications such as image description and human identification. In this paper, we introduce a novel method to combine the advantages of both multi-task and curriculum learning in a visual attribute classification framework. Individual tasks are grouped based on their correlation so that two groups of strongly and weakly correlated tasks are formed. The two groups of tasks are learned in a curriculum learning setup by transferring the acquired knowledge from the strongly to the weakly correlated. The learning process within each group though, is performed in a multi-task classification setup. The proposed method learns better and converges faster than learning all the tasks in a typical multi-task learning paradigm. We demonstrate the effectiveness of our approach on the publicly available, SoBiR, VIPeR and PETA datasets and report state-of-the-art results across the board. 
\end{abstract}
\section{Introduction}
	Moments after the Boston marathon bombing, the FBI gathered almost 10TB of photos and videos, looking for a ``backpack-carrying man, wearing a white hat''. In suspect descriptions, humans tend to rely on visual attributes since (i) they can be composed in different ways to create descriptions; (ii) they are generalizable as with some fine-tuning they can be applied to recognize objects for different tasks; and (iii) they are a meaningful semantic representation of objects or humans that can be understood by both computers and humans. Given an image of a human, a question that arises is how can someone effectively predict the corresponding visual attributes? 
	
	In this work, we propose CILICIA (CurrIculum Learning multItask ClassIfication Attributes) to address the problem of visual attribute classification from images of humans. Instead of using low-level representations which would require extracting hand-crafted features, we propose a deep learning method to solve multiple binary classification tasks. CILICIA differentiates itself from the literature as: (i) it performs end-to-end learning by feeding a single ConvNet with the entire image of a human without making any assumptions about predefined connection between body parts and image regions; and (ii) it exploits the advantages of both multi-task and curriculum learning. Tasks are split into two groups based on their cross-correlation. The group of the strongly correlated attributes is learned first, and then the acquired knowledge is transferred to the second group.  
	
	\begin{figure}[t] 
		\centering
		\includegraphics[width=0.4\textwidth]{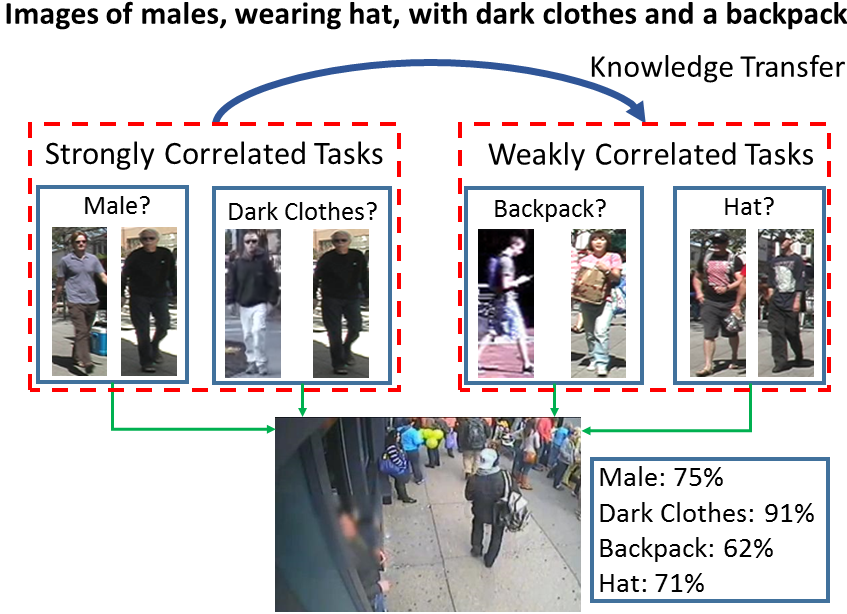}
		\caption{Can we do better in visual attribute multi-task classification? Wouldn't it be great if we could find a way to learn the attributes in a more semantically meaningful way instead of all at the same time? Our approach aspires to combine the advantages of curriculum learning and multi-task classification to predict the visual attributes of humans.} 
		\label{fig:Boston}
	\end{figure} 
	When Vapnik and Vashist introduced the learning using privileged information (LUPI) paradigm \cite{Vapnik_2009_15070}, they drew inspiration from human learning. They observed how significant the role of an intelligent teacher was in the learning process of a student, and proposed a machine learning framework to imitate this process. Employing privileged information from an intelligent teacher at training time has recently received significant attention from the scientific community with remarkable results \cite{kakadiaris2016show, motiian2016information, sarafianos2016predicting, Sharmanska_2013_16765, MVrigkas_ICIP16, Wang_2015_16756}. 
	
	Our work also draws inspiration from the way students learn in class. First, students find it difficult to learn all tasks at once. It is usually easier for them to acquire some basic knowledge first, and then build on top of that, by learning more complicated concepts. This can be achieved by learning in a hierarchical way as in the method of Yan \etal~\cite{yanhd} or with a curriculum strategy. Curriculum learning \cite{bengio2009curriculum, jiang2014self} (presenting easier examples before more complicated and learning tasks sequentially, instead of all at the same time) imitates this learning process. It has the advantage of exploiting prior knowledge to improve subsequent classification tasks but it cannot scale up to many tasks since each subsequent task has to be learned individually. However to maximize students' understanding a curriculum might not be sufficient by itself. Students also need a teaching paradigm that can guide their learning process, especially when the task to be learned is challenging. The teaching paradigm in our method is the split of visual attribute classification tasks that need to be learned into strongly and weakly correlated. In that way, we exploit the advantages of both multi-task and curriculum learning. First, the ConvNet learns the strongly correlated tasks in a multi-task learning setup, and once this process is completed, the weights of the respective tasks are used as an initialization for the more diverse tasks. During the training of the more diverse tasks, the prior knowledge obtained is leveraged to improve the classification performance. An illustrative example of our method is depicted in Figure~\ref{fig:Boston}. 
	
	In summary, this paper has the following contributions. First, we introduce CILICIA, a novel method of exploiting the advantages of both multi-task and curriculum learning by splitting tasks into two groups based on their correlation with the rest of the tasks. The tasks of each subgroup are learned in a joint manner. Thus, the proposed method learns better and converges faster than learning all the tasks in a typical multi-task learning setup. Second, we propose a scheme of transferring knowledge between the groups of tasks which reduces the convergence time and increases the performance. We performed extensive evaluations, ablation studies and an analysis of the covariates in one small-scale dataset and one medium-scale dataset and achieved state-of-the-art results. 
	
	\section{Related Work}
	\noindent\textbf{Visual Attributes}: Predicting the visual attributes of a human from an image is not a new concept as it has previously been addressed in the literature in many contexts. Ferrari and Zisserman~\cite{ferrari2007learning} were the first to investigate the power of visual attributes. They used low-level features and a probabilistic generative model to learn these attributes and segment them in an image. Kumar \etal~\cite{kumar} proposed an automatic method to perform face verification and image search by training classifiers for describable facial visual attributes (\eg, gender, hair color, and eyewear). Scheirer \etal~\cite{scheirer2012multi}  proposed a novel method to construct normalized ``multi-attribute spaces'' from raw classifier outputs. However, they focused entirely on the score calibration without investigating the feature extraction part. Following the deep learning renaissance, several papers \cite{gkioxari2015actions, rstarcnn, li2016human} have addressed the visual attribute classification
	problem using ConvNets. Zhang \etal~\cite{zhang2014panda} proposed an attribute classification method which combines part-based models in the form of poselets \cite{bourdev2011describing}, and deep learning by training pose-normalized ConvNets. Their method though, requires training a network for each poselet which is a computationally expensive task. Zhu \etal~\cite{zhu2015multi} introduced a method for pedestrian attribute classification. They proposed a ConvNet architecture comprising 15 separate subnetworks (\ie, one for each task) which are fed with images of different body parts to learn jointly the visual attributes. However, their method assumes that there is a pre-defined connection between parts and attributes, and that all tasks depend on each other and thus, learning them jointly will be beneficial. Finally, a very interesting prior work which focuses on the correlation of visual attributes is the method of Jayaraman \etal~\cite{jayaraman2014decorrelating}. While our work also leverages information from correlated attributes in a multi-task classification framework, it models co-occurrence between different groups of visual attributes instead of trying to semantically decorrelate them.
	
	\noindent\textbf{Curriculum Learning}: Solving all tasks jointly is commonly employed in the literature \cite{ciregan2012multi, hand2016attributes, zhu2015multi} as it is fast, easy to scale, and achieves good generalization. For an overview of deep multi-task learning techniques the interested reader is encouraged to refer to the work of Ruder \cite{ruder2017overview}. However, some tasks are easier than others and also not all tasks are equally related to each other \cite{pentina2015curriculum}. Curriculum Learning was initially proposed by Bengio \etal~\cite{bengio2009curriculum}. They argued that instead of employing samples at random it is better to present samples organized in a meaningful way so that less complex examples are presented first. Pentina \etal~\cite{pentina2015curriculum} introduced a curriculum learning-based approach to process multiple tasks in a sequence and developed a method to find the best order in which the tasks need to be learned. They proposed a data-dependent solution by introducing an upper-bound of the average expected error and employing an Adaptive SVM. Such a learning process has the advantage of exploiting prior knowledge to improve subsequent classification tasks but it cannot scale up to many tasks since each subsequent task has to be learned individually. 
	
	\begin{figure*}[t] 
		\centering
		\includegraphics[width=0.84\textwidth]{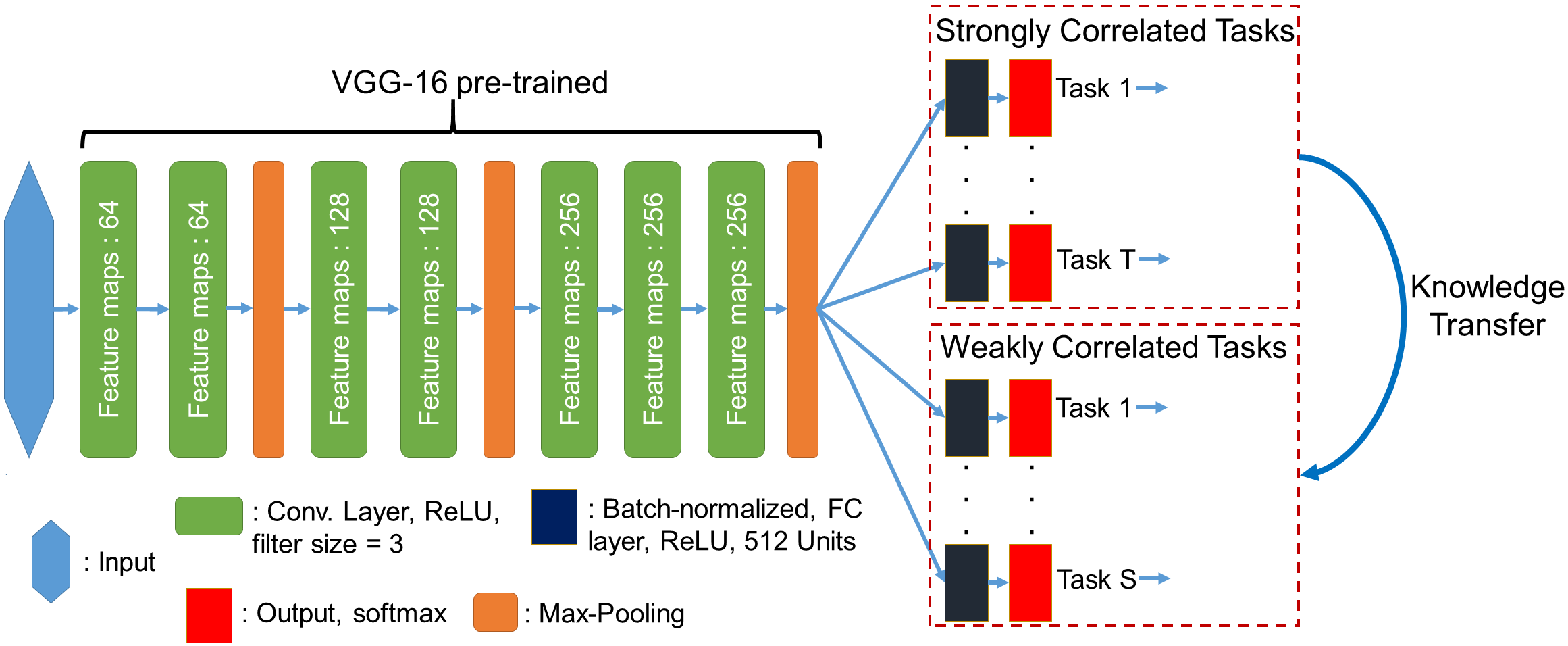}
		\caption{Architecture of the ConvNet used in our framework for both strongly and weakly correlated tasks. The VGG-16 pre-trained part is kept frozen during training and only the weights of the last layers are learned. The two parts are learned separately. However, when the weakly correlated tasks are trained, both tasks contribute to the total cost function.}
		\label{fig:Method}
	\end{figure*}
	
%
	\section{Methodology}
	In our supervised learning paradigm, we are given tuples \((x_i,y_i)\) where \(x_i\) corresponds to images and \(y_i\) to the respective visual attribute labels. The total number of tasks will be denoted by \(T\), and thus the size of \(y_i\) for one image will be \(1\times T\). Finally, we will refer to the parts of the network that solve the strongly and the weakly correlated tasks as \(C_s\) and \(C_w\), respectively. 
	
	\subsection{Multi-label ConvNet}
	To mitigate the lack of training data we employ the pre-trained VGG-16 \cite{simonyan2014very} network.  VGG-16, is the network from Simonyan and Zisserman which was one of the first methods to demonstrate that the depth of the network is a critical component for good performance. VGG-16 is trained on ImageNet \cite{russakovsky2015imagenet}, the scale of which enables us to perform transfer learning between ImageNet and our tasks of interest. The architecture of the network we use is depicted in Figure~\ref{fig:Method}. We used the first seven convolutional layers of the VGG-16 network and dropped the rest of the convolutional and fully-connected layers. The reason behind this is that the representations learned in the last layers of the network are very task dependent \cite{yosinski2014transferable} and thus, not transferable. Following that, for every task we added a batch-normalized \cite{ioffe2015batch} fully-connected layer with 512 units and a ReLU activation function. We employed batch-normalization since it enabled higher learning rates, faster convergence, and reduced overfitting. Although shuffling and normalizing each batch has proven to reduce the need of Dropout, we observed that adding a dropout layer \cite{srivastava2014dropout} was beneficial as it further reduced overfitting. The Dropout probability was 75\% for datasets with less than 1,000 training samples and 50\% for the rest. For every task, an output layer is added with a softmax activation function using the categorical cross entropy.
	
	Furthermore, we observed that the random initialization of the parameters of the last two layers backpropagated large errors in the whole network even if we used different learning rates throughout our network. To address this behavior of the network, which is thoroughly discussed in the method of Sutskever \etal~\cite{sutskever2013importance}, we ``freeze'' the weights of the pre-trained part and train only the last two layers for each task in order to learn the layer weights and the parameters of the batch-normalization. 
	
	After we ensured that we can always overfit on the training set, which means that our network is deep enough and discriminative enough for the tasks of interest, our primary goal was to reduce overfitting. Towards this direction, we (i) selected 512 units for the fully connected layer to prevent the network from learning several weights; (ii) employed a small weight decay of \(0.0001\) for the layers that are trained; (iii) initialized the learning rate at 0.001 and reduced it by a factor of 5 every 100 epochs and up to five times in total; and (iv) augmented the data by performing random scaling up to 150\% of the initial image followed by random crops, horizontal flips and adding noise by applying PCA to the RGB pixel values as proposed by Krizhevsky \etal~\cite{krizhevsky2012imagenet}. At test time, we averaged the predictions at three different scales (100\%, 125\% and 150\%) of five fixed crops and their horizontal flips (30 in total) to obtain the predicted class label. This technique, which was also adopted in the ResNet method of He \etal \cite{resNets2016}, proved to be very effective as it reduced the variation on the predictions.
	
	
	\subsection{Correlation-based Group Split}
	Finding the order in which tasks need to be learned so as to achieve the best performance is difficult and computationally expensive. Given some tasks \(t_i, i=1...T\) that need to be performed, we seek to find the best order in which the tasks should be performed so the average error of the tasks is minimized: 
	\begin{equation}\label{eq:1}
	\displaystyle \underset{\substack{S(t_i)}}{\text{minimize}} \; {\frac{1}{T}\sum\limits_{j=1}^{T} \mathcal{E}(\hat{y}_{t_j},y_{t_j}),}\\
	\end{equation}
	where \(S(t_i)\) is the function that finds the sequence of the tasks, \(\hat{y}_{t_j},y_{t_j}\) are the prediction and target vectors for task \(j\), and \(\mathcal{E}\) the prediction error. 
	
	However, the fact that a task can be easily performed does not imply that it is  positively correlated with another and that by transferring knowledge the performance of the latter will increase. Adjeroh \etal~\cite{D_2010_11241} studied the correlation between various anthropometric features and demonstrated that some correlation clusters can be derived in human metrology, whereby measurements in a cluster tend to be highly correlated with each other but not with the others. The correlation between different sub-problems was also exploited in the age estimation method of Niu \etal~\cite{Niu_2016_CVPR} in an ordinal regression setup. 
		
	To address this problem we propose to find the total dependency \(p_{i}\) of task \(t_{i}\) with the rest, by computing the respective Pearson correlation coefficients:  
	\begin{equation} \label{eu_eqnP}
	p_{i} =\sum_{j=1, j\neq i}^{T} \frac{cov(y_{t_{i}},y_{t_{j}})}{\sigma(y_{{t_{i}}})\sigma(y_{{t_{j}}})}, \; i=1,...,T
	\end{equation}
	where \(\sigma(y_{{t_{i}}})\) is the standard deviation of the labels \(y\) of the task \(t_{i}\). After we compute the total dependencies for all attributes, the obtained vector of size \(T \times 1\) (each value corresponds to one line of the Pearson correlation coefficient matrix) is sorted in a descending order. 
	Tasks with a top 50\% of \(p_i\) are strongly correlated with the rest, and thus they are assigned to the strongly correlated group. The remaining tasks are assigned as weakly correlated and will employ the information learned from the former group.  
	
	\subsection{Multi-Task Curriculum Learning}
	In the scenario we are investigating, we solve multiple binary unbalanced classification tasks simultaneously. Thus, similar to Zhu \etal~\cite{zhu2016multi} we employ the categorical cross-entropy function between predictions and targets, which for a single attribute \(t\) is defined as follows: 
	\begin{equation} \label{eu_eqn1}
	L_t = -\frac{1}{N}\sum_{i=1}^{N}\sum_{j=1}^{M} \Bigg( \frac{\nicefrac{1}{M_j}}{\sum_{n=1}^{M}\nicefrac{1}{M_n}}\Bigg)\cdot \mathds{1}[y_{i}=j] \cdot log(p_{i,j}),
	\end{equation}
	where \(\mathds{1}[y_{i}=j]\) is equal to one when the ground truth of sample \(i\) belongs to class \(j\), and zero otherwise, \(p_{i,j}\) is the respective prediction which is the output of the softmax nonlinearity of sample \(i\) for class \(j\) and the term inside the parenthesis is a balancing parameter required due to imbalanced data. The total number of samples belonging to class \(j\) is denoted by \(M_j\), \(N\) is the number of samples and \(M\) the number of classes. 
	
	\begin{algorithm}[t]
		\SetKwInOut{Input}{Input}
		\SetKwInOut{Output}{Output}
		\Input{Training set \(X\), training labels \(Y\)}
		\(Y_s, Y_w\leftarrow\) using the observations \(X\), split labels \(Y\) by maximizing Eq. (\ref{eu_eqnP})\\
		\(C_s \leftarrow\) freeze \(C_w\), train model using \((X, Y_s)\) by minimizing the loss in Eq. (\ref{eu_eqn1})  \\
		Initialize \(C_w\) from \(C_s\)\\
		\(C_w \leftarrow\) train model using \((X, Y_w)\) by minimizing the loss in Eq. (\ref{eu_eqn3})\\
		\Output{Parameters of networks \(C_s\) and \(C_w\) for the strongly and the weakly correlated tasks, respectively}
		\caption{Multi-task curriculum learning training}
		\label{alg1} 
	\end{algorithm}
	
	However, in the method of Zhu \etal~\cite{zhu2016multi} the total loss over all attributes is defined as \(L_s = \sum_{t=1}^{T}\lambda_t \cdot L_t\), where \(\lambda_t\) is the contribution weight of each parameter. For simplicity, it is set to \(\lambda_t = \nicefrac{1}{T}\), but this is problematic since there is an underlying assumption that all tasks contribute equally to the multi-task classification problem. To overcome this limitation, a fully-connected layer with \(T\) units could be added with an identity activation function after each separate loss \(L_t\) is computed. In that way, the respective weight for each attribute in the total loss function could be learned. However, we observed that for groups of tasks that consist of a few attributes the difference in the performance was statistically insignificant, and thus we did not investigate this any further. 
		
	Once the classification of the visual-attribute tasks that demonstrated a strong correlation with the rest is performed, we use the learned parameters (\ie, weights, biases and batch normalization parameters) to initialize the network for the less diverse attributes. Its architecture remains the same, with the parameters of VGG-16 being kept ``frozen''. When the number of tasks is odd, then an additional ``branch'' is added at the end of the network to learn the task-specific parameters. Furthermore, by adopting the ``supervision transfer'' technique of Zhang \etal~\cite{zhang2016real} we leverage the knowledge learned by backpropagating the following loss: 
	\begin{equation} \label{eu_eqn3}
	L_w = \lambda\cdot L_s + (1-\lambda)\cdot L_w^f,
	\end{equation}
	where \(L_w^f\) is the total loss computed during the forward pass using Eq. (\ref{eu_eqn1}) only over the weakly correlated tasks and \(\lambda\) is a parameter that controls the amount of knowledge transferred. Throughout our experimental investigation we found that a 25\% contribution of the already learned group of strongly correlated tasks yielded the best results. 
	
	The process of computing the two groups of attributes is performed once before the training starts. Since it only requires the training labels of the tasks to compute the cross-correlations and perform the split, it is not computationally intensive. Finally, note that, the group split depends on the training set and it's possible that different train-test splits might yield different groups of tasks which is why average classification results are reported over five random splits. 
	
	
	\section{Experiments}    
	\begin{table}[t]
		\centering
		\caption{Classification accuracy of different learning paradigms on the SoBiR dataset. In individual learning, each attribute is learned separately. In multi-task learning, the average loss of all attributes is backpropagated in the network. Attributes are in descending order based on their cross-correlation. Those in the second group correspond to the weakly correlated.}
		\small    
		\begin{tabular}{l c P{1.2cm} P{1.4cm} P{1.2cm}}
			\toprule
			Soft Label  & SVM & Individual Learning & Multi-Task Learning & CILICIA\\
			\midrule
			Weight & 57.7 & 67.7 & 71.0 & \textbf{73.6} \\
			Figure &  57.8 & 68.7 & 68.6 & \textbf{71.8}  \\
			Muscle build & 58.5 & 73.3 & \textbf{74.5} & 73.6 \\
			Arm thickness & 60.1 & 72.0 & \textbf{73.1} & 70.7 \\
			Leg thickness & 56.7 & 68.9 & 71.0 & \textbf{73.0} \\
			Chest size &  58.7 & 64.9 & 68.9 & \textbf{70.7} \\
			\midrule
			Age &  58.5 & \textbf{62.6} & 61.9 & 59.7 \\
			Height &  64.7 & 73.9 & 72.0 & \textbf{75.7} \\            
			Skin color & 59.2 & 66.8 & \textbf{68.0} & 67.8 \\
			Hair color &  67.5 & 74.2 & 78.1 & \textbf{78.5} \\
			Hair length & 71.8 & 78.9 & 79.2 & \textbf{79.6} \\
			Gender & 72.1 & \textbf{81.4} & 79.6 & 81.3 \\
			\midrule
			Strongly Cor.& 58.3 & 69.3 & 71.3 & \textbf{72.3} \\
			Weakly Cor.& 65.6 & 73.0 & 73.2 & \textbf{73.7} \\
			Total Av. &  61.9 & 71.2 & 72.3 & \textbf{73.1}\\
			\bottomrule
		\end{tabular}%
		\label{tab:SoBiR}%
	\end{table}%

	\subsection{Results on SoBiR}
\begin{figure}[t] 
		\centering
		\includegraphics[width=0.455\textwidth]{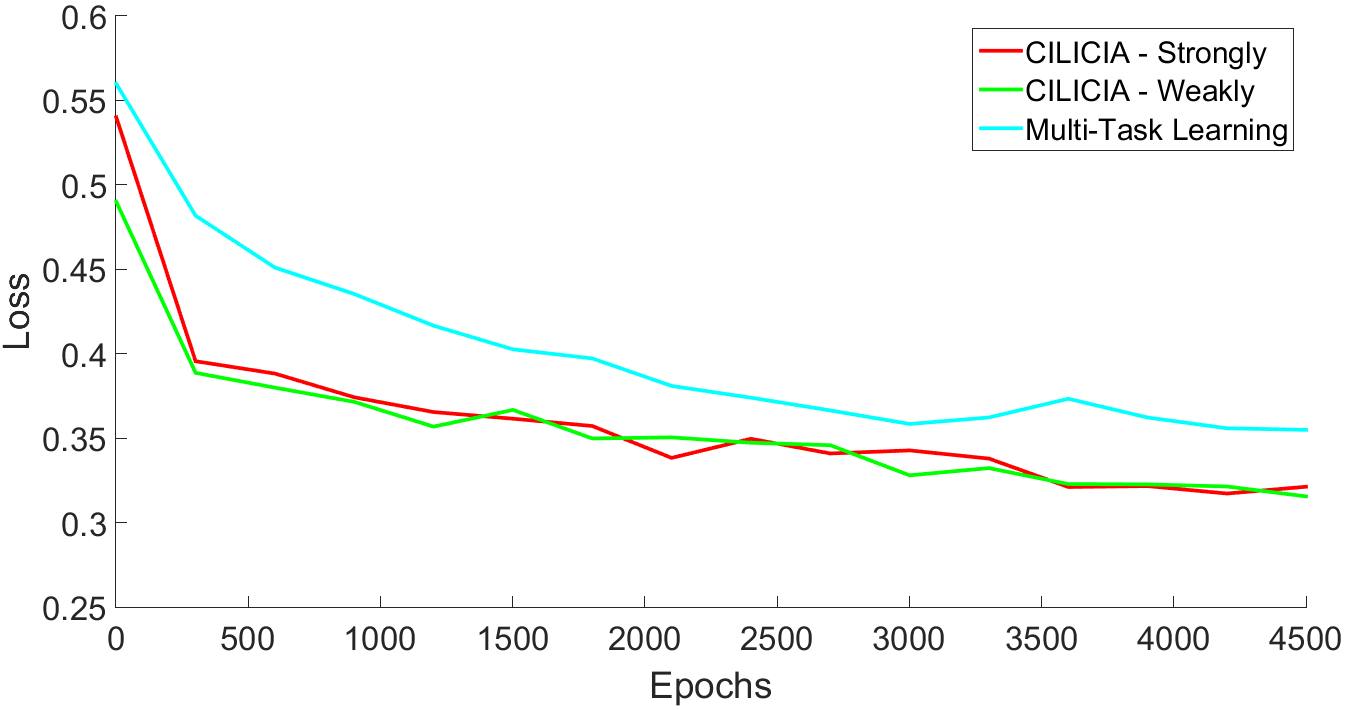}
		\caption{Convergence plot for both groups of CILICIA and Multi-Task learning on the SoBiR dataset. Note that the first group corresponds to the strongly correlated and the second to the weakly correlated group of tasks.}
		\label{fig:converg}
	\end{figure}

	Since the SoBiR dataset \cite{martinho2016soft} does not have a baseline on attribute classification we reported results using handcrafted features and an SVM classifier as well as three different end-to-end learning frameworks using our ConvNet architecture. In all cases, images were resized to \(128\times 128\). The features used for training the SVMs consisted of: (i) edge-based features, (ii) local binary patterns (LBPs), (iii) color histograms, and (iv) histograms of oriented gradients (HOGs). To preserve local information, we computed the aforementioned features in four blocks for every image resulting in 540 features in total. Furthermore, we investigated the classification performance when tasks are learned individually (\ie, by backpropagating only their own loss in the network), jointly in a typical multi-task classification setup (\ie, by backpropagating the average of the total loss in the network), and using the proposed approach.  We report the classification accuracy (\%) for all 12 soft biometrics in Table~\ref{tab:SoBiR}. CILICIA is superior in both groups of tasks to the rest of the learning frameworks. Despite the small size of the dataset, ConvNet-based methods perform better in all tasks compared to an SVM with handcrafted features. Multi-task learning methods (\ie, multi-task and CILICIA) outperform the learning frameworks when tasks are learned independently since they leverage information from other attributes. By taking advantage of the correlation between attributes, CILICIA demonstrated higher classification performance than a typical multi-task learning scenario. However, estimating the ``age'' proved to be the most challenging task in all cases as its classification accuracy ranges from 58.5\% to 62.6\% when it is learned individually using our ConvNet architecture. Finally for completeness and to demonstrate the convergence of all learning schemes, we provide in Figure~\ref{fig:converg} the convergence plots for both CILICIA groups and Multi-Task learning.

	\begin{table}[t]
		\centering
		\caption{Performance comparison on the VIPeR dataset. Attributes are in descending order based on their cross-correlation. Those in the second group correspond to the weakly correlated.}
		\footnotesize    
		\begin{tabular}{p{2cm} P{1.8cm} P{1.7cm} c}
			\toprule
			Visual Attribute & Multi-Task Learning & Zhu \etal~\cite{zhu2015multi} & CILICIA\\
			\midrule
			barelegs & 79.6 $\pm$ 0.8 &  \textbf{84.1} $\pm$ 1.1 & 82.9 $\pm$ 0.7\\
			shorts& 76.8 $\pm$ 1.1 & 81.7 $\pm$ 1.3 & \textbf{85.2} $\pm$ 0.3\\
			nocoats& 74.3 $\pm$ 1.3 &71.3 $\pm$ 0.8 & \textbf{71.3} $\pm$ 0.5 \\
			skirt& 67.2 $\pm$ 3.7 & 78.1 $\pm$ 3.5 & \textbf{86.2} $\pm$ 3.8 \\
			nolightdarkjeanscolor& 87.1 $\pm$ 1.6 & 90.7 $\pm$ 2.0 & \textbf{96.7} $\pm$ 0.4\\
			redshirt& 79.2 $\pm$ 1.9 & 91.9 $\pm$ 1.0 & \textbf{95.1} $\pm$ 0.4 \\
			patterned& 67.4 $\pm$ 3.5 & 57.9 $\pm$ 9.2 & \textbf{77.5} $\pm$ 4.3 \\
			hashandbag& 66.9 $\pm$ 3.1 & 42.0 $\pm$ 6.5 & \textbf{81.5} $\pm$ 2.7\\
			greenshirt& 70.3 $\pm$ 2.4 & 75.9 $\pm$ 5.9 & \textbf{90.5} $\pm$  2.3\\
			lightshirt& 79.5 $\pm$ 0.9 & 83.0 $\pm$ 1.2 & \textbf{84.0} $\pm$ 0.8\\
			\midrule
			blueshirt& 69.9 $\pm$ 1.7 & 69.1 $\pm$ 3.3 & \textbf{90.2} $\pm$ 0.7 \\
			lightbottoms& \textbf{79.0} $\pm$ 1.0  & \textbf{76.4} $\pm$ 1.2 & 72.5 $\pm$ 0.4 \\
			hassatchel& 72.5 $\pm$ 0.8 & 57.8 $\pm$ 2.7 & \textbf{72.8} $\pm$ 0.3 \\
			midhair& 74.3 $\pm$ 1.3 & 76.1 $\pm$ 1.8 & \textbf{77.6} $\pm$ 1.4\\
			male& 71.5 $\pm$ 1.9 & 69.6 $\pm$ 2.6  & \textbf{71.5} $\pm$ 1.2\\
			darkhair& 70.1 $\pm$ 2.0 & \textbf{73.1} $\pm$ 2.1 & 64.9 $\pm$ 1.2\\            
			hasbackpack& 68.4 $\pm$ 1.4 &64.9 $\pm$ 1.2 & \textbf{70.2} $\pm$ 0.4\\
			darkbottoms& 68.1 $\pm$ 0.9  &\textbf{78.4} $\pm$ 0.7 & 75.2 $\pm$ 0.8\\
			jeans& 74.9 $\pm$ 0.7 & \textbf{77.5} $\pm$  0.6 & 74.9 $\pm$ 0.6\\
			darkshirt& 71.0 $\pm$ 1.4 & 82.3 $\pm$ 1.4 & \textbf{84.3} $\pm$ 0.5\\
			\midrule
			Strongly Cor. Av. & 73.4  $\pm$ 2.6  & 75.7 $\pm$ 3.2 & \textbf{85.1} $\pm$ 1.0 \\
			Weakly Cor. Av. & 71.9 $\pm$ 1.8 & 72.5 $\pm$ 1.7 &  \textbf{74.8} $\pm$ 0.5 \\
			Total Av. & 73.2 $\pm$ 1.2 & 74.1 $\pm$ 1.0 & \textbf{80.5} $\pm$ 0.7 \\
			\bottomrule
		\end{tabular}%
		\label{tab:VIPeR}%
	\end{table}%

	\subsection{Results on VIPeR}
	To demonstrate the superiority of the proposed approach over normal multi-task learning approaches, we evaluate in Table~\ref{tab:VIPeR} its performance in comparison with the method of Zhu \etal~\cite{zhu2015multi} and a typical multi-task learning framework using the VIPeR dataset \cite{gray2007evaluating}. Employing the proposed multi-task curriculum learning approach is beneficial for the classification of visual attributes, as it outperformed the previous state-of-the-art by improving the total results by \(6.4\%\). Our method is superior in both groups but especially in the strongly correlated group of labels, in which the improvement is almost \(10\%\). CILICIA achieved better results in most of the tasks, which demonstrates the efficacy of our method over traditional multi-task learning approaches. The reason for this is that when some tasks are completely unrelated then multi-task learning has a negative effect as it forces the network to learn representations that explain everything, which is not possible. Additionally, we observed that color attributes tend to achieve higher performance compared to other attributes. The reason for this is that such attributes are highly imbalanced (sometimes more than one to nine) due to the way annotation is provided (\eg, when the question is ``is the human wearing a red t-shirt or not'' the answer is mainly negative).
	
	
	
	\section{Performance Analysis and Ablation Studies}
	The proposed approach outperformed the state-of-the-art in all three datasets. We argue that the main reasons for this are: (i) we exploited the correlation between different attributes and learned a model to classify them in two steps; (ii) the knowledge transfer from the strongly correlated to the weakly correlated attributes which improved the performance and reduced the required training time; and (iii) the use of a pre-trained deep architecture with the first layers frozen which was not the case in the method of Zhu \etal \cite{zhu2015multi}. To assess the impact of both contributions and to demonstrate their effectiveness we conducted two ablation studies. We selected the four most correlated and the four least correlated attributes of the PETA dataset so as to form the two groups of strongly and weakly correlated attributes. 
	
	\noindent\textbf{Effectiveness of knowledge transfer}: In the first ablation study we compare the classification accuracy of the selected tasks with and without knowledge transfer. When no knowledge is transferred we are simply training two multi-task classification frameworks. We report the obtained results in the last two columns of Table~\ref{tab:Abl1}. Transferring knowledge from the strongly to the weakly correlated group of tasks improves the performance of the latter by 1.89\% compared to a typical multi-task classification learning framework. 
	
	\noindent\textbf{Effectiveness of correlation-based split}: In the second study, we use the same eight selected attributes but instead of grouping them based on their cross-correlation, we randomly assign them to two groups. We follow exactly the same two-stage process (\ie, learning one group first and transferring knowledge to the second which is learned right after) and report the obtained results in the first column of Table~\ref{tab:Abl1}. We observe that learning in correlation-based groups of tasks is beneficial as CILICIA with and without knowledge transfer performs better than learning at random. Additionally, transferring knowledge between attributes that do not co-occur (or they are semantically completely different) has an adverse effect on the performance.
	
	\begin{table}[t]
		\centering
		\caption{Ablation experiments to assess the effectiveness of knowledge transfer and correlation-based split using the four most and the four least correlated attributes of the PETA dataset. In the random split column, the strongly and weakly groups refer only to the learning sequence as the split is not based on the correlation. CILICIA (w/o kt) refers to learning in correlation-split groups, but without knowledge transfer.}
		\small    
		\begin{tabular}{l c c c}
			\toprule
			Group  & Random Split & CILICIA (w/o kt) & CILICIA\\
			\midrule
			Strongly & 65.36 & 76.01 & 76.01 \\
			Weakly & 63.08 & 69.91 & \textbf{71.80}\\
			Total & 64.22 & 72.95 & \textbf{73.91}\\
			\bottomrule
		\end{tabular}%
		\label{tab:Abl1}%
	\end{table}%
	
	\section{Conclusion}
	In this paper, we introduced CILICIA, a multi-task curriculum learning method to address the visual-attribute classification problem. Given images of humans as an input, we performed end-to-end learning by solving multiple binary classification problems simultaneously. Tasks were grouped based on their cross-correlation so that two groups of strongly and weakly correlated tasks are formed. The attributes of each group are then learned in a multi-task learning setup. During training of the weakly correlated tasks, we leveraged the knowledge already learned from the strongly correlated tasks. By these means, we combined the advantages of both multi-task and curriculum learning paradigms; since our method converges fast, it is effective and employs prior knowledge. The obtained results demonstrate the effectiveness and, at the same time, the great potential of multi-task curriculum learning. 
	\subsection*{Acknowledgments}
	\noindent{This work has been funded in part by the UH Hugh Roy and Lillie Cranz Cullen Endowment Fund. The work of C. Nikou is supported by the European Commission (H2020-MSCA-IF-2014), under grant agreement No 656094. All statements of fact, opinion or conclusions contained herein are those of the authors and should not be construed as representing the official views or policies of the sponsors.}
	\clearpage
	{\small
		\bibliographystyle{ieee}
		\bibliography{Refs}
	}
	\clearpage

\end{document}